AI and Ethics
https://doi.org/10.1007/s43681-022-00176-2

ORIGINAL RESEARCH

# Review of the state of the art in autonomous artificial intelligence

Petar Radanliev[1] · David De Roure[1]Received: 5 March 2022 / Accepted: 16 May 2022
© The Author(s) 2022**Abstract**
This article presents a new design for autonomous artificial intelligence (AI), based on the state-of-the-art algorithms, and describes a new autonomous AI system called 'AutoAI'. The methodology is used to assemble the design founded on self-improved algorithms that use new and emerging sources of data (NEFD). The objective of the article is to conceptualise the design of a novel AutoAI algorithm. The conceptual approach is used to advance into building new and improved algorithms. The article integrates and consolidates the findings from existing literature and advances the AutoAI design into (1) using new and emerging sources of data for teaching and training AI algorithms and (2) enabling AI algorithms to use automated tools for training new and improved algorithms. This approach is going beyond the state-of-the-art in AI algorithms and suggests a design that enables autonomous algorithms to self-optimise and self-adapt, and on a higher level, be capable to self-procreate.

**Keywords** Artificial intelligence · Autonomous systems · New and emerging forms of data · AI algorithms conceptual design## 1 Introduction

The topic of artificial intelligence (AI) becoming autonomous has been discussed since the 1960s. This article reviews the current state-of-the-art in autonomous AI, with a specific focus on data preparation, feature engineering and automatic hyperparameter optimisation. Hence, this article reviews and synthesises literature and knowledge from the last decade and beyond. Using synthesised knowledge form the reviewed studies, the article presents multiple algorithms and tools in a conceptual design, to provide a new solution for automating these problems. The conceptual design uses existing AutoML techniques as the baseline for automating and assembling AI algorithms, resulting with AutoAI design superior to current AutoML. To build the AutoAI, modern AI tools can be used in combination with new and emerging forms of data (NEFD). The targets of automation are set to autonomous: data preparation, feature engineering, hyperparameter optimisation, and model selection for pipeline optimisation. The design is addressing the most important challenge in the future development and application of novel compact and more efficient AI algorithms—namely, the self-procreation of AI systems. The design consists of four Phases, each Phase addressing a number of specific obstacles (O). The design initiates with constructing training scenarios in Phase 1 that will teach the AI algorithm to use the OSINT (Big Data) for automatic ingestion of raw data from new and emerging forms of miscellaneous data formats. In Phase 2 the iterative method is be used to assemble and integrate autonomous feature selection and feature extraction—from web sites, DNS records, and OSINT sources, building upon the knowledge from autonomous data preparation and the automated feature engineering. In Phase 3 specific scenarios based on biological behaviours are designed for automatic hyperparameter optimisation at scale. In Phase 4 a new automated model selection is designed for pipeline optimisation. The design follows guidance from recent literature on fairness and ethics in AI design [1].

## 2 Methodology

This article integrates the distant fields of mathematical, computer and engineering sciences. The research has been designed with an iterative methodology, and lessons learned

✉ Petar Radanliev
  petar.radanliev@eng.ox.ac.uk

1 Department of Engineering Sciences, University of Oxford, Oxford OX1 3QG, England, UKPublished online: 07 June 2022Springer

# Review of the state of the art in autonomous artificial intelligence

Petar Radanliev[1] · David De Roure[1]

**Abstract**
This article presents a new design for autonomous artificial intelligence (AI), based on the state-of-the-art algorithms, and describes a new autonomous AI system called 'AutoAI'. The methodology is used to assemble the design founded on self-improved algorithms that use new and emerging sources of data (NEFD). The objective of the article is to conceptualise the design of a novel AutoAI algorithm. The conceptual approach is used to advance into building new and improved algorithms. The article integrates and consolidates the findings from existing literature and advances the AutoAI design into (1) using new and emerging sources of data for teaching and training AI algorithms and (2) enabling AI algorithms to use automated tools for training new and improved algorithms. This approach is going beyond the state-of-the-art in AI algorithms and suggests a design that enables autonomous algorithms to self-optimise and self-adapt, and on a higher level, be capable to self-procreate.

## 1 Introduction

The topic of artificial intelligence (AI) becoming autonomous has been discussed since the 1960s. This article reviews the current state-of-the-art in autonomous AI, with a specific focus on data preparation, feature engineering and automatic hyperparameter optimisation. Hence, this article reviews and synthesises literature and knowledge from the last decade and beyond. Using synthesised knowledge form the reviewed studies, the article presents multiple algorithms and tools in a conceptual design, to provide a new solution for automating these problems. The conceptual design uses existing AutoML techniques as the baseline for automating and assembling AI algorithms, resulting with AutoAI design superior to current AutoML. To build the AutoAI, modern AI tools can be used in combination with new and emerging forms of data (NEFD). The targets of automation are set to autonomous: data preparation, feature engineering, hyperparameter optimisation, and model selection for pipeline optimisation. The design is addressing the most important challenge in the future development and application of novel compact and more efficient AI algorithms—namely, the self-procreation of AI systems. The design consists of four Phases, each Phase addressing a number of specific obstacles (O). The design initiates with constructing training scenarios in Phase 1 that will teach the AI algorithm to use the OSINT (Big Data) for automatic ingestion of raw data from new and emerging forms of miscellaneous data formats. In Phase 2 the iterative method is be used to assemble and integrate autonomous feature selection and feature extraction—from web sites, DNS records, and OSINT sources, building upon the knowledge from autonomous data preparation and the automated feature engineering. In Phase 3 specific scenarios based on biological behaviours are designed for automatic hyperparameter optimisation at scale. In Phase 4 a new automated model selection is designed for pipeline optimisation. The design follows guidance from recent literature on fairness and ethics in AI design [1].

## 2 Methodology

This article integrates the distant fields of mathematical, computer and engineering sciences. The research has been designed with an iterative methodology, and lessons learned

✉ Petar Radanliev
  petar.radanliev@eng.ox.ac.uk

1 Department of Engineering Sciences, University of Oxford, Oxford OX1 3QG, England, UK

from each iteration are used in the design and control of the next iteration cycle—referred to as 'phases'. To reduce and overcome the complexity of the iterative methodological process, and to progress towards a better understanding of the outcome of each iteration, a variety of complimentary but different techniques are used. For example, the article intersects methodologies from engineering and computer sciences, to resolve future concerns with autonomous processing and analysing of real-time data from edge devices.

## 3 Design for autonomous AI—AutoAI

The design consists of 4 Phases. In Phase 1 automatic preparation and ingestion of raw OSINT data is synthesised for the construction of training scenarios for automated feature engineering, and to teach AI how to categorise and use (analyse) new and emerging forms of OSINT data. In Phase 2 the domain knowledge is applied to extract features from raw data. Feature is considered as valid if the attributes or properties of the feature are useful or if the characteristics are helpful to the model. For the automation of feature engineering two approaches are considered (1) multi-relational decision tree learning, which is a supervised algorithm based on decision tree and (2) Deep Feature Synthesis which is available as an open-source library named Featuretools.[1] In Phase 3 selection algorithms [2] is used to identify hyperparameter values. The normal parameters are typically optimised during training, but hyperparameters are generally optimised manually and this task is associated with a model designer. The automation scenario design with start with building upon the knowledge from biological behaviours. The particle swarm optimisation and evolutionary algorithms can be used, both of which derive from biological behaviours. The particle swarm optimisation emerges from studies on biological communities' interactions in individual and social levels and evolutionary algorithms emerge from studies on biological evolution. Second, the scenario design can apply Bayesian optimisation, which is the most used method for hyperparameter optimisation [3].

The combination of these methodological approaches is considered for automatic hyperparameter optimisation, but the concern is that edge devices are characterised by a large number of data points, and new and emerging forms of data are characterised by large configuration space and dimensionality. These factors in combination could create a longer than adequate time requirement for finding the optimal hyperparameters. Alternative method is a combined algorithm selection and hyperparameter optimisation. The method selection includes testing for the most effective approach, starting from Bayesian optimisation, Bandit Search, Evolutionary Algorithms, Hierarchical Task Networks, Probabilistic Matrix Factorisation, Reinforcement Learning and Monte Carlo Tree Search. In Phase 4 an automated pipeline optimisation is designed, comparable to the Tree-based Pipeline Optimization Tool (TPOT) [4] but for autonomous optimising feature pre-processors for maximising classification accuracy on a unsupervised classification task. The above analysis of current AI algorithms is designed for low cost devices that contain substantively larger memory than IoT sensors. In other words, this design could work for a 'Raspberry Pi' device, but it won't work for a low memory / low computation power sensor. Given this, the functionality of the proposed AutoAI needs to come in perspective. The proposed design could be applied in the metaverse, or in mobile phones, on edge devices with some memory and power, or in the metaverse. The design cannot be applied to sensors used to monitor water flow under a bridge, or the air pollution, or smoke detector sensors in the forest.

### 3.1 Phase 1: automated data preparation

$O_1$: develop open access autonomous data preparation method (for digital healthcare data) from edge devices—similar to the Oracle autonomous database,[2] for autonomous ingestion of new and emerging forms of raw data, e.g., OSINT (big data). The first scientific milestone ($M_1$) is to build a new autonomous data preparation method that can serve for training an AutoAI algorithm to: (1) become self-driving by automating the data provisioning, tuning, and scaling; (2) become self-securing by automating the data protection and security; and (3) become self-repairing by automating failure detection, failover, and repairment. The new method design includes learning how to identify, map and ignore patterns of data pollution (e.g., using direct references to results obtained from OSINT queries) and become more efficient in autonomously building improved algorithms. To ensure the success of the autonomous data preparation method, a new scenario is constructed to teach the algorithm how adversarial systems pollute the training data and how to discard such data from the training scenarios. While constructing the scenario, the search for training data expands in new and emerging forms of data (NEFD), e.g., open data—Open Data Institute,[3] Elgin,[4] DataViva[5];

---

[1] https://www.featuretools.com/.
[2] https://www.oracle.com/autonomous-database/?fbclid=IwAR2NiqrmjTZ76hj0gNa1gQUixCLEWY4g4tlvScYK0fvlW6q8HiXM-QXeC2A.
[3] https://theodi.org/.
[4] https://www.elgintech.com/.
[5] http://dataviva.info/en/.

spatiotemporal data—GeoBrick [5], Urban Flow prediction [6], Air quality [7], GIS platform [8]; high-dimensional data—Industrial big data [9], IGA–ELM [10], MDS [11], TMAP [12]; time-stamped data—Qubit,[6] Edge MWN [13], Mobi-IoST [14], Edge DHT analytics [15]; real-time data—CUSUM [16], and big data [17]. The NEFD are needed to teach the AI how to use Spark to aggregate, process and analyse the OSINT big data and to process data in RAM using Resilient Distributed Data set (RDD). NEFD can also be used to teach the AI how to use Spark Core for scheduling, optimisations, RDD abstraction, and to connect to the correct filesystem (e.g., HDFS, S3, RDBMs). With NEFD we can also train the algorithm how to use data sources from existing libraries, such as MLLib for machine learning and GraphX for graph problems. The discussed NEFD is limited to specific problems, and the AutoAI would need to be tested on that specific problem. While this research study is focused on the research stage of the work, the testing and verification stage would fail under the development stage. The development stage will require a company with a real-world use case for the AutoAI, and that could lead into new product or service development.

### 3.2 Phase 2: automated feature engineering.

$O_2$: construct automated feature engineering training scenarios to teach the AutoAI how to categorise and use (analyse) OSINT (big data) and prepare raw healthcare data for automatic ingestion. Then build training scenarios using modern tools, e.g., Recon-ng, Maltego, TheHarvester, Buscador, in combination with the prepared OSINT sources—to advance into teaching the algorithm how to autonomously build improved and transferable automated feature engineering. The fourth scientific milestone $M_4$ is to automate the feature engineering. This is a critical step in building self-procreating AI, because the performance of an algorithm depends on the quality of the import features [18]. Manual feature engineering is usually performed by an expert, e.g., data scientist with a very time-consuming trial-and-error method. First, representation learning will be applied for creating automated data pipelines from unstructured data. Representation learning is different from automated feature engineering [19], but has been proven effective in representing data for clinical predictive modelling [20]. Second, the 'expand-reduce' technique [21] will be applied for obtaining feature transformations, followed by feature selection and hyperparameter tuning. To address the known problems with the compositions of functions and the performance bottleneck, recommendations from latter updated models will be used. This will include recommendations from the ExploreKit [22], the AutoLearn [23] and their open source implementations. In addition, recommendations and experimental results will be adapted from the open-source implementation of the 'expand-reduce' techniques, e.g., Featuretools[7] and FeatureHub [24]. If these approaches fail, an alternative approach known as 'genetic programming' which is an evolutionary algorithmic technique, will be applied to mitigate the risk of failure in automating the feature engineering. The 'genetic programming' approach will be used to encode an artificial 'chromosome', then evaluate the fitness with predefined tasks and work on improving the performance. Similar approach has been used for feature engineering using a tree-based representation for feature construction and feature selection [25]. In some experiments, this approach has shown better results than the 'expand-reduce' method in terms of speed, but the solutions are unstable, because overfitting appears repeatedly. If this approach also fails, the alternative approaches that will be applied include (1) hierarchical organisation of transformations, e.g., Cognito [26] which is an automated feature engineering approach with supervised learning; (2) meta learning [27] which seem quite promising, because it uses less computational resources and better results than most other approaches; and (3) reinforcement learning with a transformation graph [28]. $M_5$: building upon the autonomous data preparation and the automated feature engineering, identify how AI can use web sites, DNS records, and OSINT sources for automated feature selection and feature extraction. Then integrate the feature selection and extraction in the training data categorisation. $M_6$: to integrate the autonomous data preparation, apply binary classification with 'dichotomization' for transferring continuous functions, variables, and equations into discrete counterparts, making them suitable for numerical evaluation though discrete mathematics (discretization). Integrate the autonomous data preparation and the automated feature engineering by applying statistical binary classification methods to the new and emerging forms of data. The statistical binary classification methods to be tested (applied) include decision trees, random forests, Bayesian networks, support vector machines, neural networks, logistic regression and a probit model. Expected results include identifying and mapping the best classifiers for a particular new and emerging form of data, e.g., random forest might perform better than support vector machines for high-dimensional data. If this approach fails, alternative methods to reach this milestone the include (1) regression analysis, e.g., Poisson regression; (2) cluster analysis, e.g., connectivity-based clustering, centroid-based clustering, distribution-based clustering, density-based clustering, grid-based clustering; and (3) learning to rank, e.g., reinforcement learning, feature vectors. $M_7$: to automate the feature selection and extraction, the methodology will: (1) use Samsara (a Scala-backed DSL language that allows for in-memory and algebraic operations) to write improved algorithms autonomously; (2) use Spark with its machine learning library; (3) use MLLib for iterative machine learning applications in-memory; (4) use MLLib for classification and regression, and to build machine-learning pipelines with hyperparameter tuning; and (5) use Samsara in combination with Mahout to perform clustering, classification, and batch-based collaborative filtering. In summary, to advance the AI algorithm into autonomously building improved and transferable automated feature engineering, first a set of modern tools (listed above) will be applied. Second, the expand-reduce feature engineering techniques will be tested and improved with the following methods: deep feature synthesis [21], ExploreKit [22], AutoLearn [23], genetic programming feature construction [25], Cognito [26], reinforcement learning feature engineering [28] and learning feature engineering [27]. While deep learning seems to be all the rage at present, for the specific problem in this study, and for enabling AI on low memory devices in general, the reinforcement learning approach seems to be a more realistic solution.

---

[6] https://www.qubit.com/.

[7] https://github.com/alteryx/featuretools.

### 3.3 Phase 3: Automated hyperparameter optimisation.

$O_3$: devise a self-optimising automated hyperparameter optimisation. Hyperparameter is a parameter that defines the values used to control the learning process. $M_8$: build a self-optimising automated hyperparameter optimisation using the particle swarm optimisation method [29], combined with evolutionary computation, e.g., deap[8] and Bayesian optimisation for building a probabilistic surrogate model, e.g., Gaussian process [30] or a tree-based model [31, 32] for automated hyperparameter optimisation. The Bayesian optimisation scenarios will be based on Python libraries, e.g., Hyperopt [33] and optimised through experimenting with black box parameter tuning, e.g., Google Vizier [34] and sequential model-based optimisation [32]. Additional sources for building the training scenario include open-source code, such as Spearmint,[9] Hyperas[10] and Talos.[11] The anticipated problems at this stage include challenges caused by configuration space, dimensionality, and the increasing number of data points. To resolve these challenges, the scenario construction will utilise the progressive sampling techniques developed for hyperparameter optimisation in large biomedical data [35] and bandit search strategy that uses successive halving [36]. If this approach also fails, an alternative approach would be to use DEvol[12] for basic proof of concept on genetic architecture search in Keras. If this approach also fails, then the next step will be to test for the best performing open-source algorithm.[13] When crafting this stage of the design, much of the effort was placed on alternatives in case the chosen approach fails. This is a crucial phase of the AutoAI, and different real-world applications will require different parameters. Hence, this phase provides great flexibility in terms of alternative approaches that are compliant with the design process.

### 3.4 Phase 4: automated model selection (and compression) for pipeline optimisation

$O_4$: build autonomous pipeline optimisers that can handle different tasks.

$M_8$: adapt different autonomous pipeline optimisers and construct different training scenarios for inference and evaluate their performance through selection consistency. Then select the most efficient pipeline optimisation model that can be applied on edge devices. The first model to use in the scenarios (and test the performance) is Auto-WEKA [37], because it is based on data mining platform and it introduced the combined algorithm selection and hyperparameter optimization (CASH) problem. The second model to use is Auto-sklearn [38] and the updated versions, e.g., PoSH Auto-sklearn [39], because of the meta-learning that improves the performance and efficacy of the Bayesian optimisation [40]. The third model to use in the scenarios is the Tree-based Pipeline Optimisation Tool (TPOT) [41], because it can perform feature pre-processing, model selection, and hyperparameter optimisation, but it has shown to make the pipelines to become overfit on the data. The testing scenarios will provide insights on how to combine the Auto-WEKA, Auto-sklearn and TPOT models for creating an efficient automated pipeline optimisation on edge devices. If these scenarios do not result with sufficient insights, additional scenarios will be created with the TuPAQ system [42] using a bandit search for optimisation based on the data and the computational budget. Followed by testing scenarios based on Auto-Tuned Models [43] which is a distributed approach based on hybrid Bayesian and/or bandit optimisation. For resolving the CASH problem, the 'ensemble learning' models, e.g., Automatic Frankensteining [44] will be used for building, selecting and ensemble of already optimised models, then increasing their prediction performance, and simultaneously reducing their input space. The ML-Plan [45] will be applied for algorithm selection and algorithm configuration, in combination with evolutionary algorithmic approaches with ensemble learning, e.g., Autostacker [46] for faster performance. For model discovery, the reinforcement learning approach will be used for the pipeline optimisation with sequence modelling method that uses deep neural networks and Monte Carlo tree searches, e.g., AlphaD3M [47]. This approach will provide new insights on the optimisation with regression and classification with faster computation times and magnitudes. In addition, the probabilistic matrix factorization [48] can be applied for predicting the pipeline performance with 'collaborative filtering', similarly to recommender systems. In summary, (1) the Auto-WEKA method can be applied with Bayesian Optimisation algorithm, (2) the Auto-Sklearn method with Joint Bayesian Optimisation and Bandit Search algorithm, (3) the TPOT method with Evolutionary Algorithms, (4) TuPAQ with Bandit Search algorithms, (5) Auto-Tuned Models with Joint Bayesian Optimisation and Bandit Search algorithms, (6) the Automatic Frankensteining with Bayesian Optimisation algorithms, (7) the ML-Plan with Hierarchical Task Networks, (8) Autostacker with Evolutionary Algorithms, (9) AlphaD3M with Reinforcement Learning/Monte Carlo Tree Search, and (10) Collaborative Filtering with Probabilistic Matrix Factorisation.

---

[8] https://github.com/DEAP/deap.
[9] https://github.com/HIPS/Spearmint.
[10] https://github.com/maxpumperla/hyperas.
[11] https://github.com/autonomio/talos.
[12] https://github.com/joeddav/devol.
[13] https://en.wikipedia.org/wiki/Hyperparameter_optimization#Open-source_software.

Table 1  Synthesising the phase 4 of the AutoAI design

| Conceptual design for a self-procreating AI | |
|---|---|
| Scientific obstacles (O) and new AI algorithms (A) | A |
| Develop a new method for autonomous data preparation and ingestion of raw data from edge devices. The method will serve for training a new AutoAI algorithm to become self-driving, self-securing and self-repairing | $O_1$ $A_1$ |
| Teach the new AutoAI how to use modern tools for scheduling, optimisations, abstraction, and to connect to correct filesystems | $O_2$ |
| Train the AutoAI algorithm how to use data sources from existing ML libraries | $O_3$ |
| Test and benchmark the efficiency and power consumption of the AutoAI | $O_4$ |
| Automate the feature engineering of AutoAI and teach the AutoAI how to categorise and analyse big data to enable the design of a self-procreating AI | $O_5$ $A_2$ |
| Develop self-procreating AutoAI neural networks based on compact representations, that can operate with lower memory requirements | $O_6$ $A_3$ |
| Identify how AI can use raw data sources for automated feature selection and feature extraction. Then integrate the feature selection and extraction in the training data categorisation | $O_7$ $A_4$ |
| Identify and map the best classifiers for a particular new and emerging form of data | $O_8$ |
| Automate the feature selection and extraction to advance the AutoAI algorithm into autonomously building improved and transferable automated feature engineering | $O_9$ $A_5$ |
| Build a self-optimising automated hyperparameter optimisation using the particle swarm optimisation method | $A_6$ |
| Build autonomous pipeline optimisers that can handle different tasks and select the most efficient model that can be applied on edge devices | $A_7$ |
| Construct tools and mechanisms for preventing bias in AI algorithms, e.g., use of less biased/more inclusive data | $O_{10}$ |

The phase 4 is summarised in as a summary map in Table 1, and it represents the final stage of the AutoAI design. The summary map outlines in great detail the step by step process for building the AutoAI. The proposed design is one step closer to a standardiser approach for AutoAI design. Such design can be transferred from resolving one problem to a completely different problem, with a relative ease. However, highly skilled AI specialists would still be required. This confirms that advancements in AI would not necessarily result with less employment opportunities in our society. It seems more likely that advancements in AI will trigger the need for re-training the workforce with new technical skills.

## 4 Discussion

Although the 'phases' are grounded on tested and verified algorithms that can enhance the autonomy of AI systems in real world scenarios, the AutoAI design is based on a set of assumptions. The scientific and technical assumptions and challenges include: (1) the development of novel AutoAI algorithms and constructing testing scenarios requires an integrated multidisciplinary multi-method approach. Using knowledge from engineering sciences, statistics and mathematics, computer science and healthcare. (2) To construct scenarios for studying novel form of AutoAI means to study a subject that does not yet exist and can take many different forms or shapes. This presents experimental and modelling challenges, while at the same time, many aspects of this research (e.g., building more compact and efficient AI) require strong experimental foundation on which the theoretical algorithmic developments can be built and validated. (3) The complexities and specificities of training the new AI algorithm with reinforced learning creates a staggering number of unpredictable parameters that need compensating. In the process of targeting these complexities with unsupervised learning and self-adaptive AI algorithms, the risk from AI itself is becoming a concern—triggering cybersecurity issues that need to be addressed. (4) The novel AutoAI algorithms represents an experimental development that will need to be tested and verified in real-world scenarios. Case study scenarios need to be constructed for testing and improving the new algorithm. The new algorithm needs to be tested for resolving problems in a controlled environment and learn how to adapt to real-life conditions. The proposed scenarios to be constructed need to be specifically targeted for problems that are interrelated and require adaptive algorithms. In other words, the case study scenarios need to be interrelated.

### 4.1 Future directions

One potential future direction for autonomous AI systems is the metaverse. A detailed review of recent literature was conducted on google scholar and majority of the articles identified are published as preprints that never reached a journal publication stage. This is interesting and intriguing, because there is a clear sign of interest on this evolution of AI in the metaverse, but there is not much in terms of strong research studies on this topic. Considering that google scholar is only one of the available search engines, alternative data reciprocities were reviewed. The next search was on the Web of Science Core Collection, and included a very simple search parameters: 'metaverse' and artificial intelligence'. This resulted with only 12 records and upon closer investigation of each record individually, it was concluded that none of the records is related to the metaverse, but instead, the records were related to 'multi-verse optimisation', and 'meta heuristic algorithms'. Since the Web of Science Core Collection 'contains over 21,100 peer-reviewed, high-quality scholarly journals published worldwide', this was considered as sufficient evidence that there is not much credible research at present time, on this topic. Hence, focus was placed on real world crypto projects that are related to AI in the metaverse. Some metaverse projects that are considered as most promising in terms of integrating AI systems (e.g., decentraland, sand, ntvtk, atlas) don't have much exposure to AI systems. The metaverse infrastructure seems to be in infancy, and the proposed design for AutoAI can be really beneficial for enabling this integration of new technologies with the proposed compact and efficient AI algorithms.

## 5 Conclusions

This article engages with designing a self-evolving Autonomous AI algorithm based on compact representations, operating on low-memory IoT edge devices. The proposed methodology for designing a new version of 'AutoAI' for low memory devices is grounded on neuromorphic engineering [49]. The proposed AutoAI is superior to current AutoML, because it is based on new and emerging forms of big data to derive transferable artificial automation that resembles a self-procreating AI. The self-procreating aspect emerges through autonomous feature selection and feature extraction, automated data preparation, automated feature engineering, automated hyperparameter optimisation, and automated model selection (and compression) for pipeline optimisation. Finally, the project engages with advancing into self-optimising and self-adaptative AI though two case study training scenarios. The novel AutoAI algorithms developed in this article, need to be tested on delivering safe and highly functional real-time intelligence. The algorithm can be used to establish the baseline for trust enhancing mechanisms in autonomous digital systems. Including the AI migration to the edge for enhancing the resilience of modern networks, such as 5G and IoT systems. The article presents a new design of a self-optimising and self-adaptive AutoAI.

### 5.1 Limitations and further research

The limitation of the proposed design is primarily in the area of limited functionality. While the proposed design can work for one AI function, it might not be as successful in all functions. Hence, the transference of code from one function to a completely different function, is something that needs to be further investigated in the trials and testing stages. Other expected difficulties include the lack of training data for unsupervised learning, and the limitations of supervised learning in terms of letting the AI algorithm learn and train itself by 'exploration' and 'exploitation'. Alternatives to mitigate this risk include (a) using reinforcement learning to develop artificial general intelligence that has the capacity to understand or learn any intellectual tasks. This will mean a shift in focus from supervised learning algorithms, such as neural networks and pattern recognition. (b) Developing stochastic process based on Bayesian and Casual statistics, instead of the current state-of-the-art strategies that are geometric based.

**Acknowledgements** Eternal gratitude to the Fulbright Visiting Scholar Project.

**Author contributions** PR: conceptualization, data curation, formal analysis, investigation, methodology, visualization, writing—original draft and review and editing. DDeR: funding acquisition, methodology, project administration, resources, software, supervision, validation, writing—review and editing.

**Funding** This work was supported by the UK EPSRC [grant number: EP/S035362/1] and by the Cisco Research Centre [grant number CG1525381].

**Availability of data and materials**  All data and materials included in the article.

## Declarations

**Conflict of interest**  On behalf of all authors, the corresponding author states that there is no conflict of (or competing) interest.

**Open Access**  This article is licensed under a Creative Commons Attribution 4.0 International License, which permits use, sharing, adaptation, distribution and reproduction in any medium or format, as long as you give appropriate credit to the original author(s) and the source, provide a link to the Creative Commons licence, and indicate if changes were made. The images or other third party material in this article are included in the article's Creative Commons licence, unless indicated otherwise in a credit line to the material. If material is not included in the article's Creative Commons licence and your intended use is not permitted by statutory regulation or exceeds the permitted use, you will need to obtain permission directly from the copyright holder. To view a copy of this licence, visit http://creativecommons.org/licenses/by/4.0/.